\documentclass[
]{ceurart}

\usepackage{listings}
\usepackage{inputenc}
\usepackage{enumitem}
\usepackage{todonotes}
\usepackage{float}
\usepackage{booktabs}

\begin{document}

\copyrightyear{2022}
\copyrightclause{Copyright for this paper by its authors.
  Use permitted under Creative Commons License Attribution 4.0
  International (CC BY 4.0).}

\conference{MediaEval'22: Multimedia Evaluation Workshop,
  January 13--15, 2023, Bergen, Norway and Online}

\title{Diffusing Surrogate Dreams of Video Scenes to Predict Video Memorability}
\renewcommand{\shorttitle}{Predicting Video Memorability}
\renewcommand{\shortauthors}{ et al.}

\author[1]{Lorin Sweeney}[
    orcid=0000-0002-3427-1250,
    email=lorin.sweeney8@mail.dcu.ie,
]\cormark[1]
\author[1]{Graham Healy}
[%
    orcid=0000-0001-6429-6339,
    email=Graham.Healy@dcu.ie
]
\author[1]{Alan F. Smeaton}
[%
    orcid=0000-0003-1028-8389,
    email=alan.smeaton@dcu.ie
]
\address[1]{Insight Centre for Data Analytics, Dublin City University, Ireland}
\cortext[1]{Corresponding author.}

\begin{abstract}
As part of the MediaEval 2022 Predicting Video Memorability task we explore the relationship between visual memorability, the visual representation that characterises it, and the underlying concept portrayed by that visual representation. We achieve state-of-the-art memorability prediction performance with a model trained and tested exclusively on \textit{surrogate dream} images, elevating concepts to the status of a cornerstone memorability feature, and finding strong evidence to suggest that the intrinsic memorability of visual content can be distilled to its underlying concept or meaning irrespective of its specific visual representational.
\end{abstract}

\maketitle
\section{Introduction and Related Work}\label{sec:intro}
The natural world is a tempest of sensory threads---from frenzied photons to odious odourants. As we wade through this storm of complex multi-sensory data, our brain is court master and king---tying threads into an intelligible internal representation, and exiling all that it deems unnecessary. What should be remembered, and what should not? The answer is hidden in the whims of the king. Memorability---the likelihood that a given piece of content will be recognised upon subsequent viewing---can be viewed as the \textit{Rosetta Stone} required to decipher the remembering whims of the brain, which is what ultimately motivates and brings meaning to its exploration. Additionally, its proximity to the essence of human experience, and ``what the brain deems to be important", casts it into the territory of proxy measure of human importance and quintessential media metric.

Although much progress has been made thinning the query-saturated haze that conceals the landscape of answers mapped by the seminal question: ``What makes an image memorable?'' \cite{image2011memorable, mem10k, sweeney2020leveraging, sweeney2021influence}, the summit remains out of sight, with 25\% of the variance still remaining unaccounted for \cite{photograph2013memorable}. The shortest path to understanding is through a hurricane of light. Given that we are visually dominant creatures, with over half of the cortex involved in visual processing \cite{snowden2012basic}, we naturally expect visual sensory data to exert the greatest influence on memorability. However, it is important not to be lead awry by our brain's appetite for visual sensory data, as semantic meaning is known to play a critical role in visual memorability. Richer and more conceptually distinctive events last longer in memory, and certain semantic categories are inherently more memorable than others \cite{photograph2013memorable, konkle2010conceptual}.
Even though visual memories are stored with an exceptional fidelity of detail (i.e., configurations and contexts of viewed objects \cite{brady2008visual}), our performance is poor when it comes to remembering random patterns unless they take on object-like qualities \cite{wiseman1974perceptual}, suggesting that visual memory is not driven entirely by visual details. 
Further evidence suggests that visual data is merely a means to conceptual understanding, which is in turn intimately tied to memory, with conceptual distinctiveness supporting higher fidelity visual long-term memory representations than perceptual distinctiveness, and influencing memory retention in a manner that cannot be accounted for by perceptual distinctiveness alone \cite{huebner2012conceptual, konkle2010conceptual}. Perceptual distinctiveness is typically measured within a given object category, and with reference to variations in low dimensional, knowledge agnostic, perceptual features (i.e., colour, and shape). Unfortunately, the line between perceptual and conceptual features begins to blur as we move into higher dimensional features (e.g., length of torso relative to head size), which become more category specific and likely to be acquired through visual experience \cite{schyns1998development}, making it difficult to probe the depth of connection between concept and memorability. However, with the recent explosion in progress in the image synthesis field, and the release of open-source text-to-image diffusion model \textit{Stable Diffusion} \cite{rombach2022high}, we find ourselves uniquely positioned to assess the impact of conceptual features on video memorability independent of its perceptual features, with the exceptional ability to preserve the depth and richness of information inherent to the visual domain.

We hypothesise that if visual data truly is merely a means to conceptual understanding, and that it is the concept itself---which is conveyed/represented through the visual data---that holds the content's intrinsic memorability, then the inter-video relationship of memorability scores predicted with ground-truth video frames should be observable in the memorability scores predicted with synthetic images predicated on purely conceptual video data.

This paper leverages state of the art image synthesis to facilitate the exploration of our aforementioned hypothesis, which can be concisely captured as the following question: can the intrinsic memorability of visual content be distilled to its underlying concept or meaning? 


\section{Approach}\label{sec:methodology}

\begin{figure}[ht]
\centering
\includegraphics[scale=0.15]{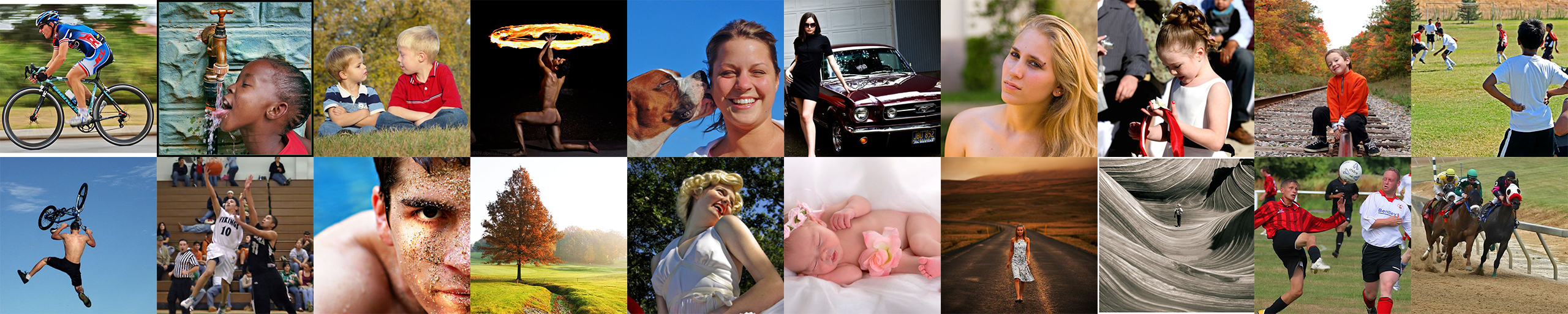}
\caption{Images used to fine-tune the Stable Diffusion model and create the mem10kstyle token.}
\label{fig:memoryTypes}
\end{figure}

\noindent Our experiments were carried out within the purview of subtask 1 of the MediaEval Predicting Video Memorability task \cite{ME2022}, with the Memento10k dataset---comprised of 7,000 training videos, 1,500 validation videos, and 1,500 withheld test videos---acting as our data landscape. However, before we could set out on our quest for insight, we had to terraform the landscape by synthesising images that reflect the \textit{conceptual essence} of the original Memento10k videos. In order to do so, we leveraged \textit{Stable Diffusion}, a latent text-to-image diffusion model \cite{rombach2022high}. Stable Diffusion is pre-trained on the LAION-5B dataset \cite{schuhmann2022laion}, which consists of scraped non-curated image-text-pairs from the internet, and is capable of generating high-resolution images from text input. 

While the images synthesised using Stable Diffusion are generally high quality in terms of image resolution, if left unspecified in the input text prompt, the compositional construction of the synthesised images is often quite unpredictable and hyper-stylised/unrealistic (i.e., cartoonish, painted, rendered). With the aim of combatting this and guiding the style of synthesised images, we created a style token (mem10kstyle) that could be appended to prompts by fine-tuning the \textit{stable-diffusion-v1-5} checkpoint on 20 real world photographs (depicted in Figure~\ref{fig:memoryTypes}.) which reflect the ``in the wild" nature of Memento10k videos, and used 1,500 Memento10k video frames as regularisation images, training for a total of 2,200 steps. 

\begin{figure}[h]
\centering
\includegraphics[scale=0.6]{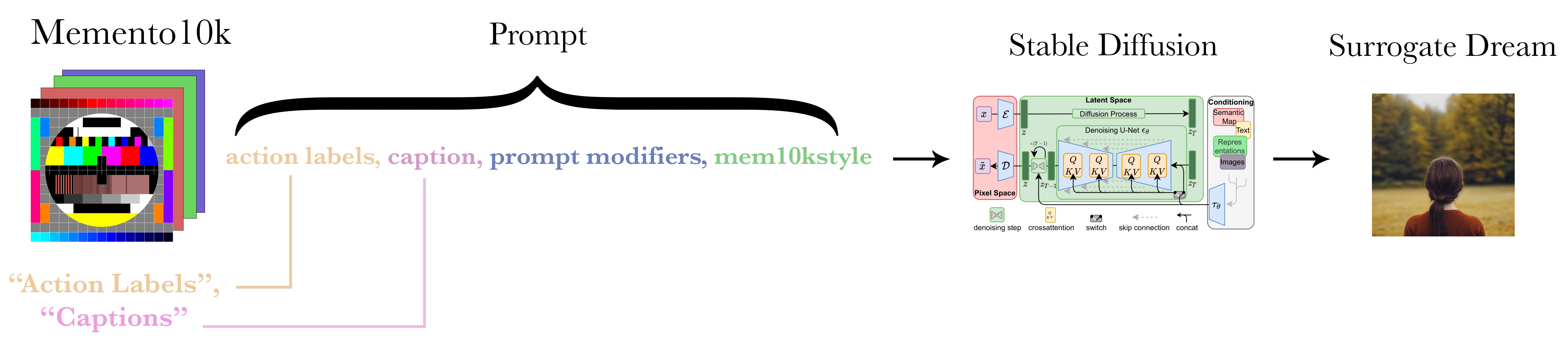}
\caption{Surrogate Dream Pipeline to synthesize images.}
\label{fig:pipeline}
\end{figure}

Stable Diffusion requires input prompts in order to generate images, so using each video's first caption as a foundation, we build a textual prompt by pre-appending video action labels, appending one of three custom prompt modifiers, and finishing with our \textit{mem10kstyle} token.
Our custom prompt modifiers are tailored to the content depicted in the video to further guide the image generation process. We then create a dataset we call \textit{``Memento10k Surrogate Dream''}---acknowledging that the synthesised images are in fact dream-like surrogates for the videos---by passing each prompt to our fine-tuned Stable Diffusion model (Figure~\ref{fig:pipeline}.)

\noindent 
We submitted 5 runs for evaluation in the Predicting Video Memorability task.  Each run falls into one of two categories: \textit{Genesis} or \textit{Surrogate Dream}.

\textbf{Genesis:} Approaches trained on vanilla Memento10k data are considered to be \textit{Genesis}, and were trained on visual features extracted from unaltered Memento10k video frames. The runs entitled {\bf Mem10k\_DenseNet121} and {\bf Mem10k\_DenseNet121\_Dream} are ImageNet-pretrained DenseNet121 models fine-tuned (for 50 epochs, with a maximum learning rate of 1e-3, and weight decay of 1e-2) on the middle frame of the Memento10k training videos. The run {\bf Mem10k\_CLIP\_Ridge\_Regression\_Mem10k} is a Bayesian Ridge Regressor (BRR) fit with default sklearn \cite{pedregosa2011scikit} parameters on stacked CLIP visual embeddings (extracted from the first, middle, and last video frames) \cite{radford2021learning}.

\textbf{Surrogate Dream:} Approaches trained on images generated with our fine-tuned Stable Diffusion model are considered to be \textit{Surrogate Dream}, and with the exception of memorability scores, were trained exclusively on surrogate visual data. The  runs entitled {\bf Dream\_DenseNet121\_Mem10k} and   {\bf Dream\_DenseNet121\_Dream} are ImageNet-pretrained DenseNet121 models fine-tuned (for 50 epochs, with a maximum learning rate of 1e-3, and weight decay of 1e-2) on our Memento10k Surrogate Dream dataset.

\section{Discussion and Outlook}
\label{sec:discussion}
%
\begin{table}[ht]
\caption{Official results on the test-set for each of our approaches.}
\label{tab:results}
\centering
\begin{tabular}{lll}
\toprule
    \textbf{Approach} & \textbf{Run Name} & \textbf{Spearman} \\
    \midrule
    \textbf{Genesis} 
    &Mem10k\_DenseNet121\_Dream & 0.583 \\
    &Mem10k\_DenseNet121\_Mem10k & 0.645 \\
    &Mem10k\_CLIP\_Ridge\_Regression\_Mem10k & \textbf{0.667} \\
    \midrule
    \textbf{Surrogate Dream} 
    &Dream\_DenseNet121\_Mem10k & 0.625 \\
    &Dream\_DenseNet121\_Dream & \textbf{0.664} \\
\bottomrule
\end{tabular}
\end{table}

\noindent Table~\ref{tab:results} shows the Spearman scores for our runs from subtask 1, with \textbf{Genesis/Surrogate Dream} indicating whether the approach was trained on ground-truth video frames, or synthesized images respectively, and the final token Mem10k/Dream of each approach indicating whether it was tested on ground-truth video frames, or 
synthesised images respectively. In the broader context of memorability prediction, all of our runs sit firmly in state-of-the-art territory, with two of our runs marginally outperforming the hitherto state-of-the-art memorability prediction model SemanticMemNet \cite{mem10k}. Although our run entitled { \bf Mem10k\_CLIP\_Ridge\_Regression\_Mem10k} achieved the highest Spearman score, the most notable aspect of our results centres around our run entitled { \bf Dream\_DenseNet121\_Dream}, which was both exclusively trained and tested on surrogate dream images, and not only outperforms our control run entitled {\bf Mem10k\_Dense121\_Mem10k}, but achieves an impressive better than state-of-the-art score of 0.664. 

\begin{figure}[h]
\centering
\includegraphics[scale=0.18]{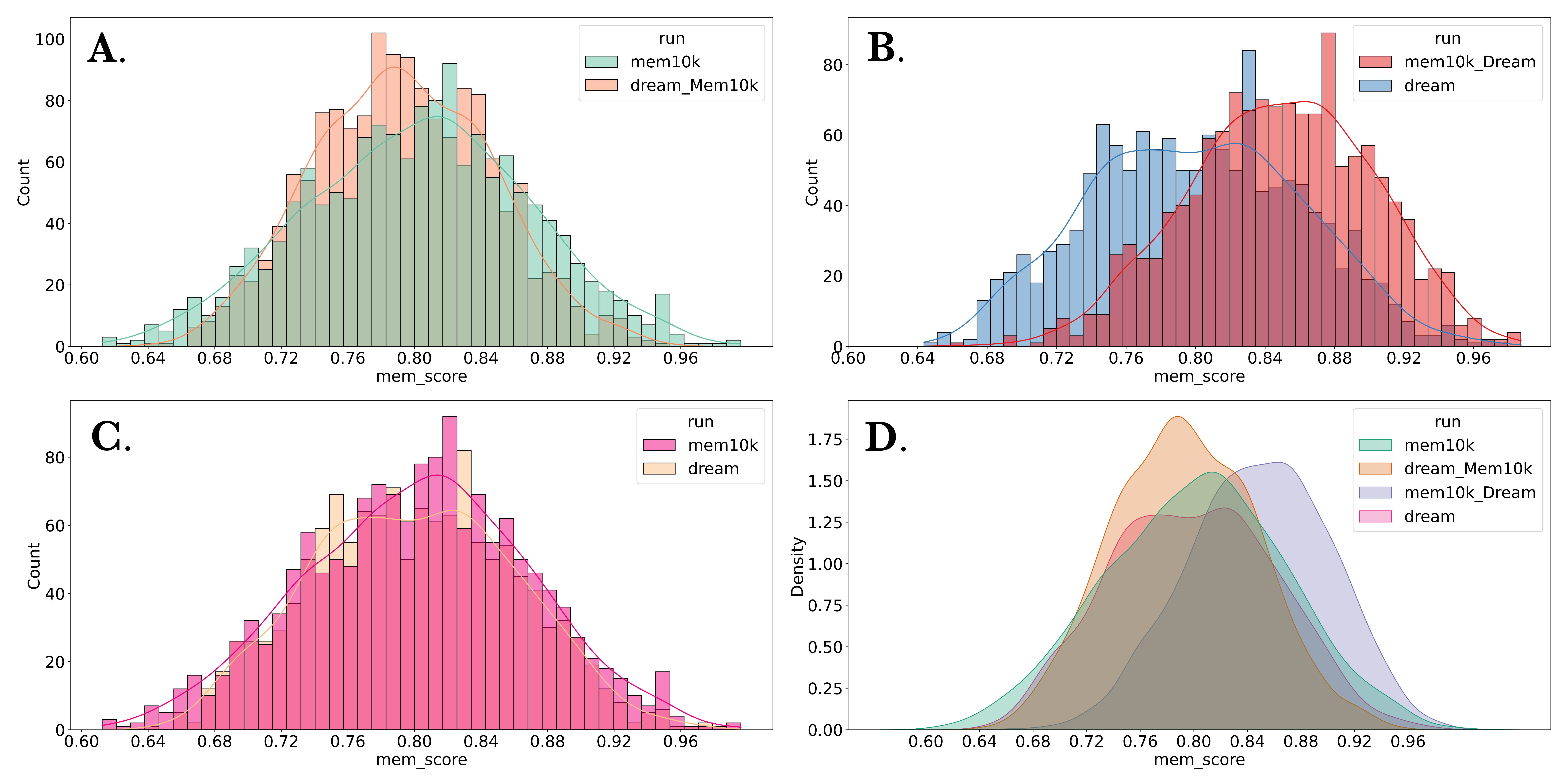}
\caption{Distribution of run predictions on official test set. Legend format: (trained on)\_(tested on).}
\label{fig:distribs}
\end{figure}

The  distributions of memorability score predictions for  vanilla and surrogate dream approaches are shown in Figure~\ref{fig:distribs}. When combined with the evaluation scores, this provides the first of its kind strong evidence that visual data is merely a means to conceptual understanding, and that it is the concepts themselves---which are conveyed/represented through the visual data---that hold the content's intrinsic memorability.

Graph B in Figure~\ref{fig:distribs} tentatively suggests that surrogate dream images are more memorable than ground-truth video frames by virtue of the left skew in predicted scores from our run trained on Mem10k frames and tested on surrogate dream images.  However, detailed exploration and investigation into the nature and composition of images in our Memento10k Surrogate Dream dataset is warranted and should be a focus of future research. 

\section*{Acknowledgements}
Science Foundation Ireland under Grant Number SFI/12/RC/2289\_P2, cofunded by the European Regional Development Fund.

\def\bibfont{\small} 
\bibliography{references} 

\end{document}